\documentclass{article}
\usepackage{spconf,amsmath,amssymb,graphicx}
\usepackage{booktabs}
\usepackage{url}
\usepackage{algorithm}
\usepackage{algpseudocode}
\let\oldtextbf\textbf\renewcommand{\textbf}[1]{\oldtextbf{\boldmath #1}}

\title{Adaptive Bidirectional Task Interaction for Joint Segmentation and
Classification of Breast Ultrasound}

\name{\begin{tabular}{@{}c@{}}
Abdullah Al Shafi$^{1,\ast}$ \qquad Md Kawsar Mahmud Khan Zunayed$^{1,\ast}$
\qquad Safin Ahmmed$^{1}$\\
Sk. Imran Hossain$^{1}$ \qquad Engelbert Mephu Nguifo$^{2}$
\end{tabular}}

\address{$^{1}$Khulna University of Engineering \& Technology, Khulna, Bangladesh\\
$^{2}$University Clermont Auvergne, Clermont Auvergne INP, ENSMSE, CNRS, LIMOS, France\\
$^{\ast}$Equal contribution}

\begin{document}
\ninept
\maketitle

\begin{abstract}
Lesion segmentation and classification in breast ultrasound inform each other,
yet many multi-task networks couple the two tasks only through shared features.
This paper introduces a decoder in which a Task Interaction Module (TIM)
exchanges latent features between the segmentation and classification branches
at each of four levels, and an Adaptive Interaction Weighting (AIW) unit uses
activation dispersion to set, per image and level, one softmax budget that the
two directions share. Multi-scale context fusion (MSCF) adds dilated context to
the residual blocks. On BUSI and BUSI-WHU, split so that duplicate images and, for
BUSI-WHU, images of one patient stay within one partition, the model exceeds
the mean IoU of CNN, transformer, multi-task and same-backbone baselines,
reaching 72.9\% and 85.4\%, and flags lesion pixels in 19.0\% of its
predictions on normal BUSI test images against 33.3\% for MTI-Net. Replacing
the per-image budget by fixed or learned constant coefficients, or by
independent sigmoid gates, costs 0.92 to 1.51 IoU points on BUSI.
Code and data partitions:
\url{https://github.com/C-loud-Nine/Adaptive-Task-Interaction-BUS}.
\end{abstract}


\section{Introduction}
\label{sec:intro}

Breast cancer is the most frequently diagnosed cancer in women
worldwide~\cite{bray2024globocan}. Ultrasound is real-time and free of ionising
radiation, but lesions vary in morphology, shadowing obscures boundaries and
speckle degrades image quality. Lesion segmentation and classification are
complementary: lesion shape and margin, which segmentation delineates, are also
cues to malignancy, and the class of a lesion constrains which shapes and
margins are plausible. Single-task models~\cite{ronneberger2015unet} solve the
two in isolation. Multi-task networks share a backbone and add task-specific
attention~\cite{liu2019endtoend,ling2024mtanet}, and in many of them the tasks
interact only through the shared features: once the decoder and the classifier
separate, neither sees what the other has computed.

Methods that couple the tasks more directly pass masks, attention maps or
distilled features between them (Section~\ref{sec:related}), with mixing that
is either fixed after training or modulated through attention maps. None of
them sets explicitly, for each image, how much of the exchange each branch
receives. The usefulness of the cross-task signal differs between images: a
well-defined lesion gives the classifier a reliable shape cue, while a shadowed
or low-contrast one may not, and class evidence from a clearly classified image
can be a better prior for the decoder than that of an ambiguous one.

This paper makes the exchange bidirectional at every decoder level and sets its
strength per image. The contributions are:
\begin{enumerate}
\item a Task Interaction Module (TIM) that exchanges features in both
directions between the segmentation decoder maps and an image-level
classification vector at each of the four decoder levels;
\item Adaptive Interaction Weighting (AIW), which splits one softmax budget
between the two directions per image and level from the activation dispersion
of the exchanged features, with a fixed normalisation at inference so that the
coefficients of a test image depend on that image alone; controls with fixed,
learned constant and independent sigmoid coefficients isolate the effect of the
shared per-image budget;
\item an evaluation on BUSI and BUSI-WHU with duplicate groups and patients kept
within one partition, released partition files, three seeds for the proposed
model, the same-backbone baseline, every ablation row and the closest
competitor, a group bootstrap with Holm correction and a cross-dataset test.
\end{enumerate}

\section{Related work}
\label{sec:related}

\textbf{Multi-task learning.} Sharing a representation between related tasks
acts as an inductive bias that can improve each of them~\cite{caruana1997multitask}.
Most multi-task networks share an encoder and attach task-specific
heads~\cite{vandenhende2022survey}; how strongly each task drives the shared
weights is then set at the loss level, by learned task
uncertainties~\cite{kendall2018multitask} or by balancing gradient
magnitudes~\cite{chen2018gradnorm}. Such weights apply to a task as a whole and
are the same for every image. Architectures that let the tasks exchange
features fall into two groups. Soft sharing mixes the activations of parallel
task networks: cross-stitch units~\cite{misra2016cross} learn one linear
combination per layer, fixed after training. Distillation designs exchange
features near the output: PAD-Net~\cite{xu2018padnet} predicts intermediate
tasks and distils them into the final ones, MTI-Net~\cite{vandenhende2020mti}
repeats this distillation at several scales and propagates it from coarse to
fine, and DenseMTL~\cite{lopes2023densemtl} exchanges features through
correlation-guided cross-task attention. These methods target dense tasks whose
features are maps of the same form; the exchange is modulated spatially by
attention or gating, but no quantity divides it between the two directions for
each image.

\textbf{Joint segmentation and classification.} MTAN~\cite{liu2019endtoend}
adds task-specific attention networks to a shared network, and
MTANet~\cite{ling2024mtanet} applies multi-task attention to medical image
segmentation and classification. In breast ultrasound,
MTL-OCA~\cite{gao2025mtloca} adds object contextual attention to the
segmentation branch and classifies from the shared backbone features. Xu et
al.~\cite{xu2022cosa} guide classification with segmentation features through
self-attention, CTG-Net~\cite{yang2022ctgnet} passes coarse masks to the
classifier and class-weighted features back to a refining segmenter, and Zheng
et al.~\cite{zheng2025mtil} combine self- and cross-task attention with circular
training. In medical imaging more broadly, mutual learning exchanges features
between the tasks under uncertainty guidance~\cite{ren2023uml}. The proposed
model differs in exchanging features at each of the four decoder levels, in
both directions, between maps and an image-level vector, and in setting the
share of the exchange per image and level.

\textbf{Data leakage in BUSI.} BUSI contains duplicate and near-duplicate
images, listed by Paw{\l}owska et al.~\cite{pawlowska2023busi}. A split at the
image level can place copies of one lesion in training and test, which inflates
test scores; splits that keep each duplicate group within one partition avoid
this~\cite{nizar2026clcyolo}. Section~\ref{ssec:data} quantifies the effect on
the partitions used here.

\section{Method}
\label{sec:method}

\begin{figure*}[t]
\centering
\includegraphics[width=\textwidth]{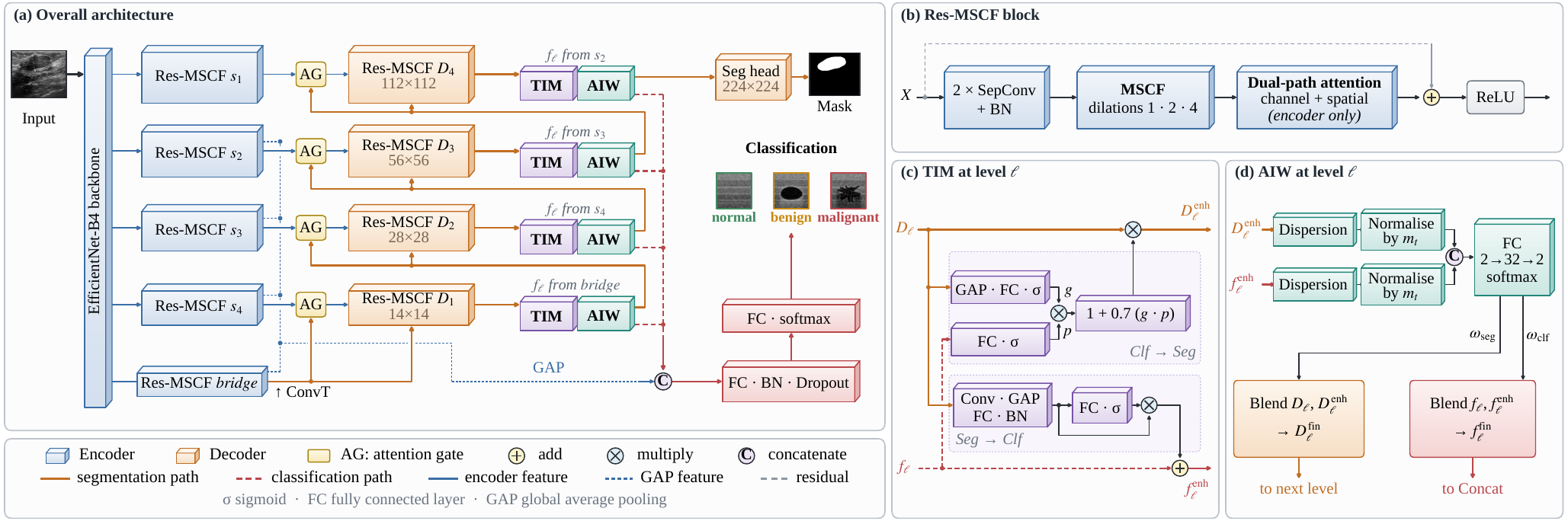}
\caption{Architecture. (a) EfficientNet-B4 encoder, four-level decoder
$D_1$--$D_4$ with attention gates (AG), and a TIM and AIW unit at each level.
(b) Residual block with MSCF. (c) TIM at level $\ell$
(Eqs.~\ref{eq:s2c}--\ref{eq:c2s}). (d) AIW at level $\ell$
(Eqs.~\ref{eq:disp}--\ref{eq:blend2}).}
\label{fig:arch}
\end{figure*}

\subsection{Network}
\label{ssec:net}

An ImageNet-pretrained EfficientNet-B4~\cite{tan2019efficientnet} encoder
feeds a four-level decoder ($D_1$--$D_4$) that restores resolution through
transposed convolutions and attention-gated~\cite{oktay2018attention} skip
connections (Fig.~\ref{fig:arch}). For a $224\times224$ input, the skip
connections $s_1$--$s_4$ are the expansion activations of the first block of
encoder stages 2, 3, 4 and 6 (144, 192, 336 and 960 channels at $112^2$,
$56^2$, $28^2$ and $14^2$), and the bridge is the final 1792-channel
$7\times7$ map. Each of these five outputs passes through a residual block with
MSCF (Section~\ref{ssec:mscf}). The decoder levels $D_1$--$D_4$ have 384, 192,
96 and 48 channels at $14^2$, $28^2$, $56^2$ and $112^2$; each upsamples the
previous level by a stride-2 transposed convolution, concatenates it with its
gated skip connection and applies a residual block with MSCF. The attention
gates add a dilated context branch (separable convolutions with rates 1 and 2)
before the sigmoid. The segmentation head upsamples $D_4$ from $112\times112$ to
the $224\times224$ mask with a transposed convolution, a $3\times3$ and a
$1\times1$ convolution. The classification head concatenates pooled encoder
outputs $s_2$ to $s_4$ and the bridge with the four blended classification
vectors $f_{\ell}^{\text{fin}}$ (Eq.~\ref{eq:blend2}), 4304 values in total,
and reduces them to 256 units with batch normalisation and dropout 0.3 before
the softmax layer. Dropout 0.3 is also applied to each decoder level after TIM
and AIW, so AIW sees the same features in training and at test time.

\subsection{Task interaction module}
\label{ssec:tim}

At decoder level $\ell \in \{1,\dots,4\}$, a classification vector
$f_{\ell} \in \mathbb{R}^{256}$ is obtained from the encoder output one level
deeper than the level's skip connection (the bridge for $D_1$ and $s_4$, $s_3$
and $s_2$ for $D_2$--$D_4$) by global average pooling (GAP) and a 256-unit ReLU
layer, so the class evidence is coarse and semantic at $D_1$ and finer at
$D_4$. Given decoder features
$D_{\ell} \in \mathbb{R}^{h_{\ell} \times w_{\ell} \times c_{\ell}}$, TIM
exchanges information in two directions. Both read the pre-interaction
features and run in parallel, so the result does not depend on the order in
which the directions are applied.

From segmentation to classification, the decoder map is reduced to
$c_{\ell}/2$ channels and pooled into a summary that updates the
classification vector through a gated residual, with BN denoting batch
normalisation:
\begin{gather}
\mathbf{s}_{\ell} = \text{BN}\big(\text{ReLU}(W_s\,\text{GAP}(\text{ReLU}(\text{Conv}_{1\times1}(D_{\ell}))))\big),
\label{eq:s2c}\\
f_{\ell}^{\text{enh}} = f_{\ell} + \sigma(W_g \mathbf{s}_{\ell}) \odot \mathbf{s}_{\ell},
\label{eq:s2c2}
\end{gather}
with $\sigma$ the logistic sigmoid and $\odot$ the element-wise product. The
summary carries what the decoder has reconstructed at this level, the extent
and arrangement of lesion activations, into the classification vector; the gate
$\sigma(W_g\mathbf{s}_{\ell})$ selects, per dimension, how much of it enters,
and the residual form leaves $f_{\ell}$ unchanged where the gate closes.

From classification to segmentation, the classification vector, a pooled
encoder descriptor, is projected to the decoder width and modulates the decoder
channels:
\begin{equation}
\mu_{\ell} = 1 + \tau\, \mathbf{g}_{\ell} \odot \mathbf{p}_{\ell},
\qquad
D_{\ell}^{\text{enh}} = D_{\ell} \odot \mu_{\ell},
\label{eq:c2s}
\end{equation}
where $\mathbf{p}_{\ell} = \sigma(W_p f_{\ell})$ and
$\mathbf{g}_{\ell} = \sigma(W_q\,\text{GAP}(D_{\ell}))$, a squeeze-and-excitation
gate~\cite{hu2018squeeze}, lie in $(0,1)^{c_{\ell}}$ and are broadcast over
spatial positions, and all $W$ are learned linear maps with bias. With
$\tau = 0.7$, $\mu_{\ell} \in (1, 1.7)^{c_{\ell}}$ amplifies each decoder
channel by up to 70\%. Because the gain is the product of
$\mathbf{p}_{\ell}$ and $\mathbf{g}_{\ell}$, a channel is amplified strongly
only when the classification vector supports it and the decoder's own
statistics mark it as active. The gain is constant over spatial positions and
never below one, so the exchange reweights decoder channels without altering
their spatial layout, suppressing a channel or changing its sign. At the four
levels TIM holds 0.44, 0.20, 0.12 and 0.09~M parameters, 0.84~M in total.

\subsection{Adaptive interaction weighting}
\label{ssec:aiw}

AIW sets, for each image and level, how much of the cross-task exchange to
retain. It summarises the enhanced features by their activation dispersion, a
parameter-free measure of how widely the level's activations spread,
\begin{equation}
v_{\text{seg}} = \frac{1}{c_{\ell}} \sum\nolimits_{c} \operatorname{Var}_{h,w}\big(D^{\text{enh}}_{\ell,c}\big),
\qquad
v_{\text{clf}} = \operatorname{Var}\big(f_{\ell}^{\text{enh}}\big).
\label{eq:disp}
\end{equation}
Because TIM rescales the decoder channels and adds a residual to the
classification vector, this dispersion combines the spread of the level's
features with the change applied by the exchange. The two descriptors differ in
scale between levels and between a map and a vector, so AIW divides each
image's dispersion $v_t$ by a mean $m_t$, which removes the overall magnitude
of the activations,
\begin{equation}
\tilde{v}_{t} = v_{t}\big/\big(m_t + \epsilon\big),
\qquad t \in \{\text{seg},\text{clf}\},
\label{eq:norm}
\end{equation}
with $\epsilon = 10^{-8}$. A normalised value above one marks an image whose
features at this level spread more than is typical. As in BN, $m_t$ is the
mean of $v_t$ over the current batch during training and a fixed value at
inference: the mean of $v_t$ over the unaugmented training images, computed in
inference mode with the weights in use, level by level from $D_1$ to $D_4$ with
the upstream levels already fixed (Algorithm~\ref{alg:aiw}). The coefficients
of a test image therefore depend on that image alone, not on the batch it is
processed in. At each level, a two-layer perceptron with weights
$W_1 \in \mathbb{R}^{32\times2}$, $W_2 \in \mathbb{R}^{2\times32}$ and biases
$b_1$, $b_2$ maps the normalised pair to two coefficients,
\begin{equation}
[\omega_{\text{seg}}, \omega_{\text{clf}}]
= \operatorname{softmax}\big(W_2\,\text{ReLU}(W_1[\tilde{v}_{\text{seg}}, \tilde{v}_{\text{clf}}]^{\top} + b_1) + b_2\big).
\label{eq:omega}
\end{equation}
The softmax gives $\omega_{\text{seg}} + \omega_{\text{clf}} = 1$, so the two
directions share one interaction budget per level: a larger coefficient for
one direction means a smaller one for the other. The outputs are
\begin{gather}
D_{\ell}^{\text{fin}} = D_{\ell} + \omega_{\text{seg}}(D_{\ell}^{\text{enh}} - D_{\ell}),
\label{eq:blend}\\
f_{\ell}^{\text{fin}} = f_{\ell} + \omega_{\text{clf}}(f_{\ell}^{\text{enh}} - f_{\ell}).
\label{eq:blend2}
\end{gather}
With Eq.~\ref{eq:c2s},
$D_{\ell}^{\text{fin}} = D_{\ell} \odot (1 + \omega_{\text{seg}}\,\tau\,\mathbf{g}_{\ell} \odot \mathbf{p}_{\ell})$,
so AIW rescales the modulation strength from $\tau$ to $\omega_{\text{seg}}\tau$
for each image, and with Eq.~\ref{eq:s2c2} it scales the gated residual added to
the classification vector by $\omega_{\text{clf}}$. At
$\omega_{\text{seg}} = 0$ a level keeps its decoder features unchanged; at
$\omega_{\text{seg}} = 1$ it adopts the full TIM output and the classification
vector keeps its own. AIW has no auxiliary objective: the coefficients receive
gradients from both task losses through Eqs.~\ref{eq:blend}--\ref{eq:blend2}.
AIW holds 648 parameters, and it and the blends are computed in single
precision under mixed-precision training.

\begin{algorithm}[t]
\caption{AIW normalisation in training and inference}
\label{alg:aiw}
\begin{algorithmic}[1]
\small
\State \textbf{Training step:} for $\ell = 1,\dots,4$, compute $v_t$ (Eq.~\ref{eq:disp}) for every image of the batch and set $m_t \gets$ batch mean of $v_t$
\State \textbf{Calibration} (after every epoch and before testing):
\For{$\ell = 1, \dots, 4$}
  \State run the network in inference mode on the unaugmented training images, with levels $1,\dots,\ell-1$ already calibrated
  \State $m_t^{(\ell)} \gets$ mean of $v_t$ at level $\ell$ over these images, $t \in \{\text{seg},\text{clf}\}$
\EndFor
\State \textbf{Inference:} use the stored $m_t^{(\ell)}$ in Eq.~\ref{eq:norm}
\end{algorithmic}
\end{algorithm}

\subsection{Multi-scale context fusion}
\label{ssec:mscf}

Lesion size varies widely within both datasets. Following atrous spatial pyramid pooling~\cite{chen2017deeplab}, multi-scale
context fusion (MSCF) applies three parallel $3\times3$ dilated separable
convolutions with rates $r \in \{1,2,4\}$, receptive fields of 3, 5 and 9
pixels, to its input $\mathbf{X}$, giving $\mathbf{F}_r$. It fuses them with a
per-image softmax over scales computed from the pooled input, a scalar
per-scale variant of selective-kernel attention~\cite{li2019selective}:
\begin{gather}
\boldsymbol{\alpha} = \operatorname{softmax}\big(W_4\,\text{ReLU}(W_3\,\text{GAP}(\mathbf{X}))\big),
\label{eq:mscf}\\
\mathbf{Y} = \text{Conv}_{1\times1}\Big(\sum\nolimits_{r} \alpha_r \mathbf{F}_{r}\Big),
\label{eq:mscf2}
\end{gather}
with $W_3$ reducing the channels by a factor of 8. The softmax makes the three
scales compete for each image, unlike channel recalibration, which reweights
channels within one scale. MSCF sits in a residual block of two separable
convolutions with BN on the five encoder outputs and at every decoder level;
encoder blocks add CBAM-style channel and spatial
attention~\cite{woo2018cbam} to its output.

\subsection{Training objective}
\label{ssec:loss}

For a batch with lesion-positive images $\mathcal{P}$, the segmentation loss is
\begin{equation}
\mathcal{L}_{\text{seg}} = \frac{1}{|\mathcal{P}|}\sum_{i\in\mathcal{P}}(1 + e^{-10\rho_i})\,L_{\text{FT},i}
+ \mathcal{L}_{\text{BCE}} + 0.15\,\mathcal{L}_{\text{edge}},
\label{eq:loss}
\end{equation}
where $L_{\text{FT},i}$ is the per-image focal Tversky
loss~\cite{abraham2019focal} (exponent 0.75, equal false-positive and
false-negative weights, i.e.\ a focal Dice loss) and the weight
$1+e^{-10\rho_i}$, with $\rho_i$ the foreground fraction of image $i$,
emphasises small lesions: it approaches 2 for a vanishing lesion and is 1.37 for
a lesion covering 10\% of the image; the term is zero when
$\mathcal{P}=\varnothing$. Computing the overlap term per image keeps
large lesions from dominating the batch. The pixel-wise binary cross-entropy
$\mathcal{L}_{\text{BCE}}$ covers all images, so normal images, whose masks are
empty, also train the segmentation branch to predict no lesion.
$\mathcal{L}_{\text{edge}}$ is the batch mean of the absolute difference between
the standard deviations of the horizontal-derivative Sobel responses of the target and
predicted masks, which penalises predicted boundaries that are sharper or
smoother than the target's. The total loss is
$0.82\,\mathcal{L}_{\text{seg}} + 0.18\,\mathcal{L}_{\text{clf}}$, with focal
cross-entropy~\cite{lin2017focal} ($\gamma=2$) as $\mathcal{L}_{\text{clf}}$.

\section{Experiments}
\label{sec:exp}

\subsection{Datasets and partitions}
\label{ssec:data}

BUSI~\cite{al2020dataset} contains 780 images (133 normal, 437 benign, 210
malignant) of 600 women aged 25 to 75, acquired in 2018 at Baheya Hospital,
Cairo, with LOGIQ E9 and LOGIQ E9 Agile scanners; lesion masks were drawn
freehand and reviewed by radiologists. BUSI-WHU~\cite{huang2025busi_whu} v3
contains 927 benign and malignant images of 816 patients aged 17 to 79,
acquired from December 2020 to December 2022 at Renmin Hospital of Wuhan
University; every image contains a lesion.

Images whose mask contradicts the label (an empty mask for a lesion or a
non-empty mask for a normal image) are discarded; none occur in either dataset.
Duplicates are grouped before splitting: BUSI images by the published duplicate
list~\cite{pawlowska2023busi}, images within each dataset whose $64\times64$
greyscale thumbnails have a Pearson correlation of 0.98 or above, and BUSI-WHU
images by patient, from the dataset's patient table. Groups whose members carry
different labels are dropped (19 BUSI and 2 BUSI-WHU images), and every other
group is kept whole within one partition, as in~\cite{nizar2026clcyolo}. In
BUSI, the 327 remaining listed images form 145 groups of 2 to 7 images, and the
thumbnail check adds no further group; the other 434 images are single. In
BUSI-WHU, 147 images belong to the 37 patients with more than one image, and
the thumbnail check joins two patients into one group. This leaves 761 BUSI
images in 579 groups and 925 BUSI-WHU images of 815 patients in 814 groups.
Each dataset is split once into 20 class-stratified folds of whole groups,
14/3/3 for training, validation and test (Table~\ref{tab:split}). The partition
files are released with the code and read by every run, so the split does not
depend on the library version.

The grouping matters for the test scores. Under a class-stratified 70/15/15
split of the same images at the image level, repeated with 1,000 random seeds,
on average 32.5\% of the BUSI test images (37 of 115) and 14.7\% of the
BUSI-WHU test images have a duplicate or an image of the same patient in the
training set; in the partitions used here none does.

\begin{table}[t]
\caption{Partitions (images per class; N, B, M: normal, benign, malignant).
BUSI-WHU patients in parentheses.}
\label{tab:split}
\centering
\fontsize{8}{9.6}\selectfont
\setlength{\tabcolsep}{3pt}
\begin{tabular}{l cccc ccc}
\toprule
& \multicolumn{4}{c}{\textbf{BUSI}} & \multicolumn{3}{c}{\textbf{BUSI-WHU}} \\
\cmidrule(lr){2-5}\cmidrule(lr){6-8}
& N & B & M & Total & B & M & Total \\
\midrule
Training   & 100 & 298 & 141 & 539 & 391 & 246 & 637 (567) \\
Validation &  10 &  66 &  31 & 107 &  84 &  61 & 145 (123) \\
Test       &  21 &  63 &  31 & 115 &  84 &  59 & 143 (125) \\
\midrule
Total      & 131 & 427 & 203 & 761 & 559 & 366 & 925 (815) \\
\bottomrule
\end{tabular}
\end{table}

\begin{table*}[t]
\caption{Comparison on BUSI and BUSI-WHU (\%; parameters in millions).
Proposed, same-backbone and MTI-Net rows: mean$_{\pm\text{std}}$ over three
seeds; other rows: single runs (seed 42). IoU, Dice: lesion-positive test
images; Acc, AUC: all test images. $^{\dagger}$Classification head added: the original has no
image-level classification output; transformers are segmentation-only. $^{\ddagger}$Ablation baseline
(Table~\ref{tab:abl}, row 1): the proposed encoder, decoder and head without
MSCF, TIM and AIW. Best in \textbf{bold}; differences in the text use unrounded
means.}
\label{tab:cmp}
\centering
\fontsize{8}{9.6}\selectfont
\setlength{\tabcolsep}{2.4pt}
\begin{tabular}{l cccc cccc c}
\toprule
& \multicolumn{4}{c}{\textbf{BUSI}} & \multicolumn{4}{c}{\textbf{BUSI-WHU}} & \\
\cmidrule(lr){2-5}\cmidrule(lr){6-9}
\textbf{Method} & \textbf{IoU} & \textbf{Dice} & \textbf{Acc} & \textbf{AUC}
& \textbf{IoU} & \textbf{Dice} & \textbf{Acc} & \textbf{AUC} & \textbf{Params} \\
\midrule
U-Net~\cite{ronneberger2015unet}$^{\dagger}$ & $61.8$ & $71.2$ & $81.7$ & $84.2$ & $76.8$ & $83.5$ & $84.6$ & $89.6$ & 31.3 \\
Attention U-Net~\cite{oktay2018attention}$^{\dagger}$ & $63.3$ & $72.8$ & $82.6$ & $85.7$ & $78.4$ & $84.8$ & $85.3$ & $90.8$ & 31.7 \\
UNet++~\cite{zhou2018unetpp}$^{\dagger}$ & $65.5$ & $74.9$ & $83.5$ & $84.8$ & $77.6$ & $83.7$ & $86.7$ & $91.3$ & 9.3 \\
\midrule
TransUNet~\cite{chen2021transunet} & $67.8$ & $77.6$ & -- & -- & $82.4$ & $88.2$ & -- & -- & 105.3 \\
Swin-Unet~\cite{cao2022swin} & $68.5$ & $78.5$ & -- & -- & $83.1$ & $88.8$ & -- & -- & 34.3 \\
MISSFormer~\cite{huang2023missformer} & $66.2$ & $76.2$ & -- & -- & $82.1$ & $88.0$ & -- & -- & 42.5 \\
\midrule
MTAN~\cite{liu2019endtoend}$^{\dagger}$ & $63.9$ & $73.5$ & $81.7$ & $85.2$ & $78.1$ & $84.5$ & $86.0$ & $90.9$ & 40.2 \\
MTANet~\cite{ling2024mtanet} & $69.4$ & $79.2$ & $87.0$ & $89.0$ & $83.4$ & $88.9$ & $89.5$ & $93.4$ & 24.8 \\
MTL-OCA~\cite{gao2025mtloca} & $67.5$ & $77.3$ & $84.3$ & $87.9$ & $81.2$ & $87.2$ & $87.4$ & $92.1$ & 8.4 \\
MTI-Net~\cite{vandenhende2020mti}$^{\dagger}$ & $71.2_{\pm 0.7}$ & $80.8_{\pm 0.6}$ & $87.8_{\pm 0.9}$ & $90.8_{\pm 0.5}$ & $83.9_{\pm 0.5}$ & $89.2_{\pm 0.5}$ & \textbf{$91.6_{\pm 0.7}$} & $94.0_{\pm 0.3}$ & 11.4 \\
\midrule
Same backbone$^{\ddagger}$ & $68.2_{\pm 1.1}$ & $78.1_{\pm 1.0}$ & $87.0_{\pm 0.9}$ & $88.3_{\pm 0.7}$ & $82.0_{\pm 0.9}$ & $87.9_{\pm 0.8}$ & $87.9_{\pm 0.4}$ & $92.4_{\pm 0.6}$ & 37.8 \\
\midrule
\textbf{Proposed} & \textbf{$72.9_{\pm 0.4}$} & \textbf{$82.4_{\pm 0.4}$} & \textbf{$90.1_{\pm 0.5}$} & \textbf{$92.6_{\pm 0.3}$} & \textbf{$85.4_{\pm 0.5}$} & \textbf{$90.7_{\pm 0.4}$} & $91.1_{\pm 0.4}$ & \textbf{$94.9_{\pm 0.3}$} & 57.3 \\
\bottomrule
\end{tabular}
\end{table*}

\subsection{Implementation}
\label{ssec:impl}

Images are resized bilinearly to $224\times224$ and masks by nearest neighbour,
then binarised at 127; multi-lesion BUSI masks are merged. Each training image
is kept with three augmented copies, generated once per seed, each horizontally
flipped with probability 0.5 and rotated within $\pm15^\circ$ with probability
0.7, with reflected borders. Vertical flips are not used, since they would move
the skin line to the bottom of the image and invert posterior acoustic
shadowing and enhancement.

Models are trained in mixed precision with Adam ($\beta_1=0.91$,
$\beta_2=0.999$), batch size 24 and gradient clipping at norm 1 for up to 100
epochs. The learning rate starts at $3\times10^{-4}$, follows a cosine decay to
3\% of that value, is multiplied by 0.55 after 7 epochs without a validation
gain (cooldown 2) and is kept above $1.5\times10^{-6}$. In the proposed model
and its ablation variants the encoder (ImageNet-1k weights trained with
AutoAugment, \texttt{tf\_efficientnet\_b4} in timm) receives images in
$[0,255]$, normalises them with its ImageNet statistics, uses stochastic depth
0.2 and is frozen through the depthwise convolution of its fourth MBConv block;
the frozen layers keep their batch normalisation statistics. Layers outside the
encoder use Glorot-uniform initialisation. Of the 57.3~M parameters, the encoder
holds 17.5~M, the residual blocks on its outputs 27.1~M, the decoder with the
mask head 9.9~M, the classification layers 1.9~M and TIM 0.84~M. The MSCF units
account for 18.7~M, of which 13.3~M sit in the 1792-channel bridge block. The
model needs 18.4~GFLOPs (9.2~G multiply-accumulates) per $224\times224$ image.

The first three epochs (at most 100 batches each) are not used for selection.
From the fourth epoch on, the validation set selects the model: after every
epoch AIW is calibrated on the training images as at test time
(Algorithm~\ref{alg:aiw}), the score, 0.7 times the foreground IoU pooled over
validation pixels plus 0.3 times the accuracy, is recorded, and the best epoch
is restored (early stopping after 22 epochs without a gain of at least 0.004).
The selected model is thus evaluated in the form in which it is tested.
Segmentation-only models are selected on the IoU term.

\textbf{Comparison methods.} The comparison methods are re-implemented in
PyTorch following their papers and official code, and trained under the same
protocol: partitions, augmentation, segmentation and classification losses,
optimiser, schedule, model selection and metrics. TransUNet (R50-ViT-B/16,
ImageNet-21k), Swin-Unet (Swin-T, ImageNet-1k), MTANet (ResNet-50, ImageNet)
and MTI-Net (HRNet-W18, ImageNet) start from pretrained encoders; the others
are trained from scratch, as published. U-Net, Attention U-Net and UNet++
receive a classification head on the pooled bottleneck (fully connected layers
of 256 and $C$ units); MTAN pools its classification attention branch into the
same head, and MTI-Net's classification task uses its task-feature and
distillation machinery with global pooling. Deep supervision is kept where the
original training defines it (MTANet and MTI-Net, weight 1; MTL-OCA soft object
regions, weight 0.4). The proposed model, the same-backbone baseline, every
ablation row and MTI-Net, the closest competitor, are trained with three seeds
(42, 43, 44), which vary initialisation, augmentation and batch order; the
other comparison methods are trained once (seed 42).

\subsection{Evaluation}
\label{ssec:eval}

For a test image with ground-truth lesion $G$ and prediction $P$ thresholded at
0.5, $\text{IoU} = |P\cap G|/|P\cup G|$ and
$\text{Dice} = 2|P\cap G|/(|P|+|G|)$; both are averaged over the 94 (BUSI) and
143 (BUSI-WHU) lesion-positive test images. Accuracy and AUC (macro
one-vs-rest on BUSI, binary on BUSI-WHU) cover all test images. On one seed, a
single test image moves accuracy by 0.87 points on BUSI and 0.70 on BUSI-WHU;
over three seeds, one prediction moves the mean by a third of that. On the 21
normal BUSI test images, whose masks are empty, the number of predictions with
any lesion pixel and the predicted lesion area as a fraction of the image,
averaged over all normal predictions, are reported instead. In a cross-dataset
test, models trained on BUSI-WHU are also tested, without adaptation, on the 630
BUSI lesion images (427 benign, 203 malignant) from another hospital, none a
near-copy of a BUSI-WHU image.

Differences in per-image IoU (averaged over seeds where there are several) are
tested with a paired bootstrap: each of 10,000 resamples draws the test
duplicate/patient groups with replacement and takes the image-level mean difference, so
copies of one lesion count as one unit. The 2.5th and 97.5th percentiles give
an unadjusted 95\% CI, and twice the smaller share of resamples on either side
of zero gives a two-sided $p$-value, Holm-corrected over the ten published
comparison methods on each dataset.

\begin{figure*}[t]
\centering
\includegraphics[width=0.85\textwidth]{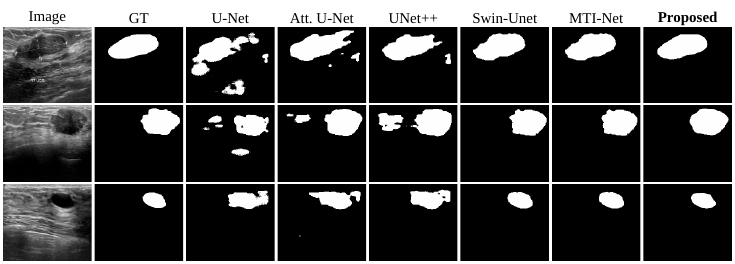}
\caption{Qualitative comparison on BUSI (seed 42). From left to right: input,
ground truth, U-Net, Attention U-Net, UNet++, Swin-Unet, MTI-Net and the
proposed model.}
\label{fig:qual}
\end{figure*}

\subsection{Comparison with existing methods}
\label{ssec:cmp}

In Table~\ref{tab:cmp}, the proposed model has the highest IoU, Dice and AUC on
both datasets and the highest accuracy on BUSI. It exceeds the same-backbone
baseline, which shares its encoder, decoder, head and training, by 4.66 (BUSI)
and 3.43 (BUSI-WHU) IoU points, and raises AUC from 88.3\% to 92.6\% and from
92.4\% to 94.9\%. The closest
competitor on both datasets is MTI-Net, which distils features between the
tasks at several scales: the proposed model is
1.71 IoU points higher on BUSI (95\% CI $-0.84$ to 4.26, $p=0.19$) and 1.56
higher on BUSI-WHU (95\% CI 0.08 to 3.04, $p=0.042$). MTI-Net is 0.47 points
higher in BUSI-WHU accuracy, two of 429 test predictions over the three seeds
(393 against 391). Across seeds the
IoU of the proposed model has a standard deviation of 0.4 (BUSI) and 0.5
(BUSI-WHU), against 0.7 and 0.5 for MTI-Net and 1.1 and 0.9 for the
same-backbone baseline.

Among the published baselines, every one that starts from a pretrained encoder
exceeds every one trained from scratch in IoU on both datasets (BUSI: 67.8 to
71.2 against 61.8 to 67.5; BUSI-WHU: 82.4 to 83.9 against 76.8 to 82.1). The
proposed model and the same-backbone baseline share their pretrained encoder,
so the 4.66 and 3.43 points between them come from MSCF, TIM and AIW. With MSCF alone
(Table~\ref{tab:abl}, row 2; 56.5~M parameters) the network reaches 69.9\% IoU on BUSI,
below MTI-Net with 11.4~M, and TIM and AIW then add 2.99 points with 0.84~M
parameters.

\textbf{Normal images.} On the 21 normal BUSI test images, 12 of the 63
predictions of the proposed model (three seeds; 19.0\%) contain lesion
pixels, against 21 of 63 (33.3\%) for MTI-Net, and the predicted lesion area averages 0.38\% of the
image against 1.12\%. In Fig.~\ref{fig:qual}, U-Net, Attention U-Net and UNet++
likewise produce spurious regions that the proposed model avoids.

\textbf{Cross-dataset test.} Trained on BUSI-WHU and tested on BUSI, the
proposed model (three-seed mean) reaches 63.4\% IoU and 76.7\% benign/malignant accuracy,
against 61.2\% and 74.4\% for MTI-Net (three-seed mean). Both lose about a
quarter of their in-domain IoU under the change of hospital and scanner; the
proposed model keeps 74.2\% of it and MTI-Net 72.9\%.

\subsection{Ablation study}
\label{ssec:abl}

\begin{table}[t]
\caption{Ablation and controls on BUSI (\%, mean$_{\pm\text{std}}$, three
seeds). Rows 1--5 add components to the same-backbone baseline; rows 6--9
replace the per-image softmax coefficients of the full model.}
\label{tab:abl}
\centering
\fontsize{8}{9.6}\selectfont
\setlength{\tabcolsep}{3pt}
\begin{tabular}{rl ccc}
\toprule
& \textbf{Configuration} & \textbf{IoU} & \textbf{Acc} & \textbf{AUC} \\
\midrule
1 & Baseline & $68.2_{\pm 1.1}$ & $87.0_{\pm 0.9}$ & $88.3_{\pm 0.7}$ \\
2 & + MSCF & $69.9_{\pm 0.8}$ & $88.1_{\pm 0.5}$ & $90.1_{\pm 0.5}$ \\
3 & + TIM & $69.4_{\pm 0.9}$ & $87.5_{\pm 0.5}$ & $89.5_{\pm 0.7}$ \\
4 & + TIM + AIW & $70.2_{\pm 0.8}$ & $88.4_{\pm 0.5}$ & $90.9_{\pm 0.4}$ \\
5 & Full (MSCF + TIM + AIW) & \textbf{$72.9_{\pm 0.4}$} & \textbf{$90.1_{\pm 0.5}$} & \textbf{$92.6_{\pm 0.3}$} \\
\midrule
6 & Full, $\omega=1$ (no AIW) & $71.3_{\pm 0.6}$ & $89.0_{\pm 0.5}$ & $91.4_{\pm 0.4}$ \\
7 & Full, $\omega=0.5$ & $71.6_{\pm 0.5}$ & $89.3_{\pm 0.5}$ & $91.8_{\pm 0.4}$ \\
8 & Full, sigmoid gates & $71.9_{\pm 0.5}$ & $89.6_{\pm 0.9}$ & $92.0_{\pm 0.4}$ \\
9 & Full, learned constant $\omega$ & $71.8_{\pm 0.5}$ & $89.3_{\pm 0.5}$ & $91.9_{\pm 0.4}$ \\
\bottomrule
\end{tabular}
\end{table}

\textbf{Components.} In Table~\ref{tab:abl}, MSCF raises the baseline IoU by
1.67 points and TIM with AIW by 1.98 (TIM alone 1.21, AIW a further 0.77), or
by 2.99 with MSCF present (row 2 to row 5). The two contributions are not
additive: together they raise the baseline IoU by 4.66 points, 1.01 more than
the sum of their separate gains, so the exchange gains most when it operates on
features with multi-scale context. Classification follows the same order: AUC
rises from 88.3 to 90.1 with MSCF, to 90.9 with TIM and AIW, and to 92.6 with
all three, and adding AIW to TIM raises AUC from 89.5 to 90.9, more than it
raises IoU.

\textbf{Controls on the coefficients.} Rows 6 to 9 keep MSCF and TIM and change
only how the exchange is weighted. Rows 6, 7 and 9 remove the per-image
coefficients, keeping the full TIM output
($\omega_{\text{seg}}=\omega_{\text{clf}}=1$), a fixed half (both 0.5) or one
learned softmax pair per level; row 9 is the counterpart, inside this network,
of mixing weights that are learned and then fixed, as in cross-stitch
units~\cite{misra2016cross}. Row 8 keeps the per-image coefficients but
replaces the softmax by independent sigmoid gates, so the two directions no
longer share one budget. The full model exceeds rows 6, 7, 8 and 9 by 1.51,
1.24, 0.92 and 1.08 IoU points, and has the highest accuracy and AUC of all
rows. Even without AIW (row 6), TIM on MSCF features adds 1.48 points to row 2.
Halving the exchange for every image (row 7) recovers 0.27 of the 1.51 points
lost without AIW, and learning one constant pair per level (row 9) adds 0.16 to
that. Per-image coefficients without a shared budget (row 8) recover 0.59, and
sharing one budget between the two directions gives the remaining 0.92.

\textbf{Behaviour of AIW.} In the full model, $\omega_{\text{seg}}$ varies
across BUSI test images with a standard deviation of 0.042 and is positively
correlated with lesion area (Spearman correlation 0.23; means over levels). Since the budget is
shared, larger lesions tend to shift the exchange towards modulating the
decoder and smaller ones towards updating the classification vector. Small as
this spread is, the best constant pair (row 9) is 1.08 IoU points below the
per-image coefficients.

\section{Discussion}
\label{sec:disc}

The controls separate three properties of the exchange. The largest step is
the exchange itself: every configuration with TIM on MSCF features (rows 5 to
9, at least 71.3\% IoU) exceeds every configuration without that combination
(rows 1 to 4, at most 70.2\%). Weighting the exchange by a constant gains
little: a fixed half or a learned constant pair recovers at most 0.43 of the
1.51 points between the unweighted exchange and the full model. The rest
requires both remaining properties together. Per-image coefficients without a
shared budget (row 8) stay 0.92 points below the full model, and a shared
budget without per-image coefficients (row 9) stays 1.08 below; only their
combination reaches 72.9\%. Rows 8 and 5 differ only in whether the two
directions share one budget: with independent gates an image can take the full
exchange in both directions at once, and the shared budget rules this out.

The interaction between MSCF and the exchange is the other structural result.
TIM and AIW add 1.98 IoU points to the baseline and 2.99 to the network with
MSCF, and the summary that TIM passes to the classifier is pooled from decoder
features whose receptive field MSCF has widened. The exchange is cheap relative
to its effect: TIM and AIW hold 1.5\% of the parameters, while MSCF, which holds
a third of them, adds 1.67 IoU points on its own.

The evaluation protocol shapes how these numbers should be compared with
published results on BUSI. Without grouping, about a third of BUSI test images
have a duplicate in training (Section~\ref{ssec:data}), so scores obtained on
image-level splits are not comparable with those reported here. All methods in
Table~\ref{tab:cmp} are trained and tested on the same released partitions.

\section{Conclusion}
\label{sec:concl}

The full model, which exchanges latent features in both directions at every
decoder level under one per-image budget, gives the highest mean IoU among the
compared methods on duplicate-aware splits of BUSI and BUSI-WHU, significantly
above MTI-Net on BUSI-WHU; against MTI-Net it also gives fewer false-positive
predictions on normal images and higher cross-dataset IoU and accuracy. TIM and
AIW together add 0.84~M parameters. Fixed or
learned constant coefficients or independent sigmoid gates in place of this
budget lose 0.92 to 1.51 IoU points on BUSI, and multi-scale context and
cross-task exchange together gain more than the sum of their separate
contributions. The released partitions, code and per-run checks allow these
comparisons to be repeated on the same images.

\section{Compliance with ethical standards}
\label{sec:ethics}

This research study was conducted retrospectively using human subject data made
available in open access by BUSI~\cite{al2020dataset} and BUSI-WHU
v3~\cite{huang2025busi_whu}. Ethical approval was not required as confirmed by
the licence attached with the open access data; the BUSI-WHU collection was
approved by the Renmin Hospital of Wuhan University ethics committee
(WDRY2022-K217).

\section{Acknowledgments}
\label{sec:ack}

No funding was received, and the authors have no conflicts of interest.
Large language model tools were used to assist with editing and accelerate
LaTeX formatting. All technical claims, derivations, and experimental results
were authored, verified, and are the responsibility of the authors.

\bibliographystyle{IEEEbib}
\bibliography{main}

@article{bray2024globocan,
  author  = {Bray, F. and others},
  title   = {Global cancer statistics 2022: {GLOBOCAN} estimates of incidence and mortality worldwide},
  journal = {CA Cancer J. Clin.}, volume = {74}, number = {3}, pages = {229--263}, year = {2024},
  doi     = {10.3322/caac.21834}
}

@article{al2020dataset,
  author  = {Al-Dhabyani, W. and Gomaa, M. and Khaled, H. and Fahmy, A.},
  title   = {Dataset of breast ultrasound images},
  journal = {Data Brief}, volume = {28}, pages = {104863}, year = {2020},
  doi     = {10.1016/j.dib.2019.104863}
}

@misc{huang2025busi_whu,
  author = {Huang, J. and others},
  title  = {{BUSI-WHU}: Breast cancer ultrasound image dataset},
  note   = {{Mendeley Data}, V3, doi: 10.17632/k6cpmwybk3.3}, year = {2025}
}

@inproceedings{liu2019endtoend,
  author    = {Liu, S. and Johns, E. and Davison, A. J.},
  title     = {End-to-end multi-task learning with attention},
  booktitle = {CVPR}, pages = {1871--1880}, year = {2019},
  doi       = {10.1109/CVPR.2019.00197}
}

@article{ling2024mtanet,
  author  = {Ling, Y. and Wang, Y. and Dai, W. and Yu, J. and Liang, P. and Kong, D.},
  title   = {{MTANet}: Multi-task attention network for automatic medical image segmentation and classification},
  journal = {IEEE Trans. Med. Imag.}, volume = {43}, number = {2}, pages = {674--685}, year = {2024},
  doi     = {10.1109/TMI.2023.3317088}
}

@article{gao2025mtloca,
  author  = {Lu, Y. and Sun, F. and Wang, J. and Yu, K.},
  title   = {Automatic joint segmentation and classification of breast ultrasound
             images via multi-task learning with object contextual attention},
  journal = {Front. Oncol.}, volume = {15}, pages = {1567577}, year = {2025},
  doi     = {10.3389/fonc.2025.1567577}
}

@inproceedings{misra2016cross,
  author    = {Misra, I. and Shrivastava, A. and Gupta, A. and Hebert, M.},
  title     = {Cross-stitch networks for multi-task learning},
  booktitle = {CVPR}, pages = {3994--4003}, year = {2016},
  doi       = {10.1109/CVPR.2016.433}
}

@inproceedings{vandenhende2020mti,
  author    = {Vandenhende, S. and Georgoulis, S. and Van Gool, L.},
  title     = {{MTI-Net}: Multi-scale task interaction networks for multi-task learning},
  booktitle = {ECCV}, pages = {527--543}, year = {2020},
  doi       = {10.1007/978-3-030-58536-5_31}
}

@inproceedings{lopes2023densemtl,
  author    = {Lopes, I. and Vu, T.-H. and de Charette, R.},
  title     = {Cross-task attention mechanism for dense multi-task learning},
  booktitle = {WACV}, pages = {2329--2338}, year = {2023}
}

@inproceedings{ronneberger2015unet,
  author    = {Ronneberger, O. and Fischer, P. and Brox, T.},
  title     = {{U-Net}: Convolutional networks for biomedical image segmentation},
  booktitle = {MICCAI}, pages = {234--241}, year = {2015},
  doi       = {10.1007/978-3-319-24574-4_28}
}

@misc{oktay2018attention,
  author  = {Oktay, O. and others},
  title   = {Attention {U-Net}: Learning where to look for the pancreas},
  howpublished = {arXiv:1804.03999}, year = {2018}
}

@inproceedings{zhou2018unetpp,
  author    = {Zhou, Z. and Siddiquee, M. M. R. and Tajbakhsh, N. and Liang, J.},
  title     = {{UNet++}: A nested {U-Net} architecture for medical image segmentation},
  booktitle = {DLMIA}, pages = {3--11}, year = {2018},
  doi       = {10.1007/978-3-030-00889-5_1}
}

@inproceedings{tan2019efficientnet,
  author    = {Tan, M. and Le, Q. V.},
  title     = {{EfficientNet}: Rethinking model scaling for convolutional neural networks},
  booktitle = {ICML}, pages = {6105--6114}, year = {2019}
}

@article{chen2017deeplab,
  author  = {Chen, L.-C. and Papandreou, G. and Kokkinos, I. and Murphy, K. and Yuille, A. L.},
  title   = {{DeepLab}: Semantic image segmentation with deep convolutional nets, atrous convolution, and fully connected {CRFs}},
  journal = {IEEE Trans. Pattern Anal. Mach. Intell.}, volume = {40}, number = {4}, pages = {834--848}, year = {2018},
  doi     = {10.1109/TPAMI.2017.2699184}
}

@inproceedings{hu2018squeeze,
  author    = {Hu, J. and Shen, L. and Sun, G.},
  title     = {Squeeze-and-excitation networks},
  booktitle = {CVPR}, pages = {7132--7141}, year = {2018},
  doi       = {10.1109/CVPR.2018.00745}
}

@inproceedings{woo2018cbam,
  author    = {Woo, S. and Park, J. and Lee, J.-Y. and Kweon, I. S.},
  title     = {{CBAM}: Convolutional block attention module},
  booktitle = {ECCV}, pages = {3--19}, year = {2018},
  doi       = {10.1007/978-3-030-01234-2_1}
}

@article{pawlowska2023busi,
  author  = {Paw{\l}owska, A. and Karwat, P. and {\.Z}o{\l}ek, N.},
  title   = {Letter to the editor. {Re: `Dataset of breast ultrasound images' by W. Al-Dhabyani et al.}},
  journal = {Data Brief}, volume = {48}, pages = {109247}, year = {2023},
  doi     = {10.1016/j.dib.2023.109247}
}

@misc{chen2021transunet,
  author  = {Chen, J. and others},
  title   = {{TransUNet}: Transformers make strong encoders for medical image segmentation},
  howpublished = {arXiv:2102.04306}, year = {2021}
}

@inproceedings{cao2022swin,
  author    = {Cao, H. and others},
  title     = {{Swin-Unet}: Unet-like pure transformer for medical image segmentation},
  booktitle = {ECCV Workshops}, pages = {205--218}, year = {2022},
  doi       = {10.1007/978-3-031-25066-8_18}
}

@article{huang2023missformer,
  author  = {Huang, X. and Deng, Z. and Li, D. and Yuan, X. and Fu, Y.},
  title   = {{MISSFormer}: An Effective Transformer for {2D} Medical Image
             Segmentation},
  journal = {IEEE Trans. Med. Imag.}, volume = {42}, number = {5}, pages = {1484--1494}, year = {2023},
  doi     = {10.1109/TMI.2022.3230943}
}

@inproceedings{lin2017focal,
  author    = {Lin, T.-Y. and Goyal, P. and Girshick, R. and He, K. and Doll{\'a}r, P.},
  title     = {Focal loss for dense object detection},
  booktitle = {ICCV}, pages = {2980--2988}, year = {2017},
  doi       = {10.1109/ICCV.2017.324}
}

@inproceedings{abraham2019focal,
  author    = {Abraham, N. and Khan, N. M.},
  title     = {A novel focal {Tversky} loss function with improved attention {U-Net} for lesion segmentation},
  booktitle = {ISBI}, pages = {683--687}, year = {2019},
  doi       = {10.1109/ISBI.2019.8759329}
}

@article{zheng2025mtil,
  author  = {Zheng, S. and Li, J. and Qiao, L. and Gao, X.},
  title   = {Multi-task interaction learning for accurate segmentation and
             classification of breast tumors in ultrasound images},
  journal = {Phys. Med. Biol.}, volume = {70}, number = {6}, pages = {065006},
  year    = {2025}, doi = {10.1088/1361-6560/adae4d}
}

@inproceedings{li2019selective,
  author    = {Li, X. and Wang, W. and Hu, X. and Yang, J.},
  title     = {Selective kernel networks},
  booktitle = {CVPR},
  pages     = {510--519},
  year      = {2019}
}

@inproceedings{xu2022cosa,
  author    = {Xu, M. and Huang, K. and Qi, X.},
  title     = {Multi-task learning with context-oriented self-attention for breast ultrasound image classification and segmentation},
  booktitle = {ISBI},
  pages     = {1--5},
  year      = {2022},
  doi       = {10.1109/ISBI52829.2022.9761685}
}

@inproceedings{ren2023uml,
  author    = {Ren, K. and others},
  title     = {Uncertainty-informed mutual learning for joint medical image classification and segmentation},
  booktitle = {MICCAI},
  pages     = {35--45},
  year      = {2023}
}

@article{yang2022ctgnet,
  author    = {Yang, K. and others},
  title     = {{CTG-Net}: Cross-task guided network for breast ultrasound diagnosis},
  journal   = {PLoS One},
  volume    = {17},
  number    = {8},
  pages     = {e0271106},
  year      = {2022}
}

@misc{nizar2026clcyolo,
  author = {Nizar, M. F. and Rachmatullah, M. N. and Supardi, J.},
  title  = {{CLC-YOLO}: A compact channel-gated prototype network for real-time leakage-aware breast ultrasound lesion segmentation},
  howpublished = {arXiv:2609.31702}, year = {2026}
}

@article{caruana1997multitask,
  author  = {Caruana, R.},
  title   = {Multitask learning},
  journal = {Mach. Learn.}, volume = {28}, number = {1}, pages = {41--75}, year = {1997},
  doi     = {10.1023/A:1007379606734}
}

@inproceedings{kendall2018multitask,
  author    = {Kendall, A. and Gal, Y. and Cipolla, R.},
  title     = {Multi-task learning using uncertainty to weigh losses for scene geometry and semantics},
  booktitle = {CVPR}, pages = {7482--7491}, year = {2018}
}

@inproceedings{chen2018gradnorm,
  author    = {Chen, Z. and Badrinarayanan, V. and Lee, C.-Y. and Rabinovich, A.},
  title     = {{GradNorm}: Gradient normalization for adaptive loss balancing in deep multitask networks},
  booktitle = {ICML}, pages = {794--803}, year = {2018}
}

@inproceedings{xu2018padnet,
  author    = {Xu, D. and Ouyang, W. and Wang, X. and Sebe, N.},
  title     = {{PAD-Net}: Multi-tasks guided prediction-and-distillation network for simultaneous depth estimation and scene parsing},
  booktitle = {CVPR}, pages = {675--684}, year = {2018}
}

@article{vandenhende2022survey,
  author  = {Vandenhende, S. and Georgoulis, S. and Van Gansbeke, W. and Proesmans, M. and Dai, D. and Van Gool, L.},
  title   = {Multi-task learning for dense prediction tasks: A survey},
  journal = {IEEE Trans. Pattern Anal. Mach. Intell.}, volume = {44}, number = {7}, pages = {3614--3633}, year = {2022},
  doi     = {10.1109/TPAMI.2021.3054719}
}

\end{document}